\documentclass{article}
\usepackage{PRIMEarxiv}
\usepackage{amsmath,graphicx,hyperref}
\usepackage{amssymb}
\RequirePackage[numbers, sort]{natbib}

\usepackage{booktabs}
\usepackage{multirow}

\usepackage{xcolor}

\def\bbx{{\mathbf x}}
\def\bbo{{\mathbf o}}
\def\bbv{{\mathbf v}}
\def\bbz{{\mathbf z}}
\def\bbD{{\mathbf D}}
\def\reals{{\mathbb{R}}}

\title{
Hyperbolic Graph Representation Learning: Embed in One Metric, Optimize with Another
}

\author{Federico Larroca, Paola Bermolen, Marcelo Fiori and Bernardo Marenco\\
  Facultad de Ingenier\'ia \\
  Universidad de la Rep\'ublica \\
  Uruguay\\
  \texttt{\{flarroca, paola, mfiori, bmarenco\}@fing.edu.uy}
  }
  
\begin{document}
\maketitle
\begin{abstract}
Hierarchical graphs embed in hyperbolic space with lower distortion than in Euclidean space owing to its negative curvature. However, their gradient-based learning is hampered at large radii, where the Poincar\'e ball and the Lorentz hyperboloid models fail numerically. Polar coordinates avoid this problem, but the hyperbolic metric scales the angular step by the hyperbolic sine of the radius, freezing angular motion. We observe that this factor is a choice, silently fixed by existing implementations: the Euclidean tangent parametrization, for instance, uses the radius itself. We show that other choices are not only possible but preferable. They are endpoints of a one-parameter family of optimization preconditioners with curvatures from $-1$ to $0$, while the embedding remains at curvature $-1$. We show that since the Euclidean preconditioner rearranges a layout but refines it poorly, while an intermediate one refines far better once a layout is in place, combining them in two stages reduces the loss on real-world trees by 46--74\% over the best single curvature.

\end{abstract}
\section{Introduction}
\label{sec:intro}

The analysis, modeling, and optimization of complex networks from the perspective of machine learning is one of the most important topics in data science today with countless applications. Many of these techniques are based on the fundamental assumption that there is a latent geometry in which the network nodes are located, and these nodes are then interpreted as vectors in that space~\cite{hamilton2020graph}. 
Traditionally, techniques for estimating these vectors (embeddings) assume an Euclidean geometry.

However, hyperbolic geometry has emerged as an alternative representation capable of capturing structures, such as hierarchical, tree-like and large-hubs networks,  with very few dimensions~\cite{krioukov2010hyperbolic,sarkar2011low}. A wide family of techniques for hyperbolic embeddings assumes knowledge of a certain distance between nodes, and the goal is to position the embeddings in space in a way that best reflects these distances. Some of these methods include Poincaré Maps \cite{klimovskaia2020poincare}, Hydra \cite{keller2020hydra}, and Poincaré Embeddings \cite{nickel2017poincare} (or their version in the Lorentz model of hyperbolic spaces, known as Lorentz Embeddings \cite{nickel2018learning}). 
To find the embeddings that best reflect these distances, both Poincaré Maps and Poincaré/Lorentz Embeddings perform gradient descent on the manifold defined by hyperbolic space~\cite{bonnabel2013stochastic,becigneul2019riemannian}. Hydra instead applies a gradient based algorithm from a good initial point calculated from spectral decompositions.

Although optimization over manifolds is well understood~\cite{Boumal_2023}, hyperbolic representations suffer from well-known numerical issues at large radii~\cite{sala2018representation,yu2019numerically,mishne23numerical}, which is precisely where hierarchical graphs place most of their nodes: both the Poincar\'e ball and the Lorentz hyperboloid lose precision through catastrophic cancellations. Polar coordinates remain accurate over a much wider range of radii, but the hyperbolic metric scales the angular component by the hyperbolic sine of the radius, which freezes angular motion under gradient-based optimization.

The central observation of this work is that the metric used to take gradient steps need not be the metric in which the embeddings live. The loss is always evaluated with the hyperbolic distance, so the optimizer's metric acts only as a preconditioner: it changes the path followed by the optimizer, not the problem being solved. Existing implementations silently fix this choice: Riemannian methods step with the exact hyperbolic metric, whereas the Euclidean tangent parametrization of~\cite{mishne23numerical} implicitly scales the angular component by the radius itself. Our contributions are threefold. (i) We show that these two choices are the endpoints of a one-parameter family of preconditioners, given by the metrics of constant curvature running from $-1$ to $0$, while the embedding space remains at curvature $-1$ throughout since the loss is untouched (unlike mixed-curvature approaches~\cite{gu2019learning}, which change the embedding space itself). (ii) We explain why both endpoints fail when nodes span a wide range of radii: the hyperbolic one freezes angular motion, and the flat one cannot move inner and outer nodes with a single learning rate. (iii) We show that other members of the family yield better results, particularly in a two-stage scheme (flat first, then an intermediate curvature) that reduces the stress on real-world trees by 46--74\% over the best single curvature.

\section{Metric and preconditioning}
\label{sec:charts}

There are many hyperbolic embedding methods, basically differing in the loss they minimize. In our experiments, and as an example, we will consider the so-called \emph{stress}~\cite{keller2020hydra}: the squared difference between the embeddings' hyperbolic distance (at fixed curvature -1) and the corresponding shortest-path distance in the graph; i.e.,
\begin{gather*}
   \mathcal{L}= \sum_{i<j}\left(d_{\mathbb{H}}(\bbx_i,\bbx_j) - \bbD_{ij}\right)^{2}
\end{gather*}
where $\bbD$ is the matrix of graph shortest-path distances and $\bbx_i$ the coordinates of the embedding of the $i$-th node.

Furthermore, several representations of hyperbolic geometry exist (and
thus ways of representing each point $\bbx$ and computing
$d_{\mathbb{H}}$), although two dominate hyperbolic representation
learning: the Poincar\'e ball~\cite{nickel2017poincare,klimovskaia2020poincare}
and the Lorentz hyperboloid~\cite{nickel2018learning}. However, both lose
precision at large radii. The Lorentz model recovers distances from a
Minkowski product of two nearly equal numbers of order $e^{2r}$, and the
cancellation already corrupts the distance of two nodes at distance 1 by
1\% at $r\approx 16$, and becomes zero at $r\approx 19$. The ball, whose points approach its boundary as $1-\tanh(r/2)\approx2e^{-r}$, suffers the same kind of cancellation.

\subsection{Polar coordinates}
An alternative to avoid these failures is to use polar coordinates~\cite{mishne23numerical}. Fixing an origin $\bbo$, a point is stored as a hyperbolic radius and a unit direction; i.e., as $\bbx = (r,\bbv)$, with $r = d_{\mathbb{H}}(\bbo,\bbx)$ and $\|\bbv\| = 1$.
This already gives a much larger usable span in the radius up to $r \approx 350$, provided the distance is evaluated in the equivalent form
$\cosh d_{\mathbb{H}}(\bbx_1,\bbx_2) = 1 + 2\sinh^2\tfrac{r_1-r_2}{2}
  + 2\sinh r_1 \sinh r_2 \sin^2\tfrac{\Delta\theta}{2}$,
with $\cos\Delta\theta = \bbv_1^\top\bbv_2$.

In what follows we work in $\mathbb{H}^2$ to keep the formulas readable and focus on vanilla Gradient Descent for ease of exposition, but everything carries over to $\mathbb{H}^d$ and other momentum-based methods (such as Adam, which we will use in the experiments). Setting
$r_1 = r_2 = r$ in the distance formula above, expanding for small $\Delta\theta$ gives
$d_{\mathbb{H}} \approx \sinh(r)\,\Delta\theta$. Since $r$ is itself an arc length along a
fixed direction and the two displacements are orthogonal, slightly changing by $(\delta r,\delta\theta)$ the coordinates of a
point at $(r,\theta)$ moves it a hyperbolic distance $\delta s$ such that
\begin{equation}\label{eq:arclength}
  \delta s^2 = \delta r^2 + \sinh^2(r)\, \delta\theta^2 .
\end{equation}
This is almost the familiar formula from plane polar coordinates, where the identical statement
reads $\delta s^2 = \delta r^2 + r^2 \delta\theta^2$; they differ in how much arc one
radian buys at radius $r$.

Equation \eqref{eq:arclength} is a quadratic cost on coordinate displacements,
$\delta s^2 = \delta\bbx^\top G_{\mathbb{H}}\,\delta\bbx$ with
$\delta\bbx = (\delta r, \delta\theta)$ and $G_{\mathbb{H}}(r) = \mathrm{diag}\left(1,\ \sinh^2 r\right)$.
The natural gradient method on $\mathbb{H}^d$ is the one that measures step size with this
cost. Among all displacements of a fixed $G$-length, the one that decreases $f$ fastest is
$\delta\bbx \propto -G^{-1}\nabla f$, so with learning rate $\eta$ the optimizer moves by
\begin{equation}\label{eq:step-exact}
  \delta r = -\eta\, \partial_r f, \qquad
  \delta\theta = -\eta\, \frac{\partial_\theta f}{\sinh^2 r}.
\end{equation}
The radial component is untouched, as it will be with every metric we
consider, so the entire question is angular. The angular derivative is
\emph{divided} by $\sinh^2 r$, and how much that costs is clearest once the coordinate step is converted back into a
displacement. By \eqref{eq:arclength} a change $\delta\theta$ moves the point along an arc
$\Delta s = \sinh(r)\,|\delta\theta|$, so under \eqref{eq:step-exact}
\begin{equation}\label{eq:arc-exact}
  \Delta s = \eta\,\frac{|\partial_\theta f|}{\sinh r},
\end{equation}
which decays exponentially with the radius. At the radii a hierarchical embedding
may reach ($r \approx 20$ or $30$ as we will see below) the angular coordinates are frozen no matter how large
$|\partial_\theta f|$ is, and the optimizer can no longer move. 

\subsection{Other metrics as preconditioner} \label{sec:preconditioner}
So far, nothing obliges the optimizer to measure its steps with
$G_{\mathbb{H}}$. The loss is evaluated using $d_\mathbb{H}$ whatever the optimizer does, so the $\sinh^2 r$ of $G_{\mathbb{H}}$ is not part of the problem being solved, it belongs to the rule that turns a gradient into a step, and we are free to change that rule. Concretely, replacing it by an arbitrary positive warp $w$ we have that $G_w(r) = \mathrm{diag}\left(1,\ w(r)^2\right)$ and $\delta s_w^2 = \delta r^2 + w(r)^2 \delta\theta^2$. Repeating the argument above results in that the step the optimizer
actually takes is now $\delta r = -\eta \partial_r f$ and $\delta\theta = -\eta\partial_\theta f/w(r)^2$ (i.e., substitute $\sinh r$ by $w(r)$ in \eqref{eq:step-exact}).

Since the difficulty is that angular motion freezes, a smaller warp is an obvious solution.
Mishne et al.~\cite{mishne23numerical} end up stepping with $w(r) = r$, although for the
representational reason we mentioned before rather than this one: they store each embedding as a single
unconstrained vector $\bbz = r\,\bbv \in \reals^d$, which ranges over all of $\reals^d$ and so
neither saturates nor overflows, and apply ordinary Euclidean gradient descent to it. Since
$\|\delta\bbz\|^2 = \delta r^2 + r^2 \delta\theta^2$ for $\bbz = r\bbv$, stepping along
$\nabla_{\bbz} f$ with $G = I$ is exactly $G_w$ with $w(r) = r$. Plain Euclidean descent on
$\bbz$ is therefore not the absence of a preconditioner but a specific choice: the warp of a \emph{flat} metric. Thus, storing coordinates and stepping along their gradient always commits one to some $w$,
whether or not a choice was intended.

We may actually interpret both proposals above (Euclidean and Hyperbolic) as the endpoints of a one-parameter family $w_c(r) = \sinh(\sqrt{c}\,r)/\sqrt{c}$, 
where $c=1$ recovers the metric of the hyperbolic space the embeddings live in and $c\to 0$ the Euclidean (or flat) 
$w_0(r)=r$. Each $w_c$ is in fact the warp of a
constant-curvature $-c$ metric, but we repeat that it acts here only as a preconditioner: the loss is untouched, the distance between embeddings is that of curvature $-1$ for every $c$, and the
warp enters only in moving the coordinates.

As we mentioned before, the flat preconditioner removes the freeze, since in this case \eqref{eq:arc-exact} reads as $\Delta s = \eta|\partial_\theta f|\sinh r/r^2$. However, if we have embeddings with radial coordinates spanning between a large $r_{\max}$ and a small $r_{\min}$, then the ratio between the movement of the outer and inner nodes scales like $e^{r_{\max}-r_{\min}}$ unless the derivative of $f$ with respect to $\theta$ decreases exponentially with $r$ (which in the case of the stress is actually the opposite). This means that, when applied on a large radial span configuration, a single learning rate cannot usefully move all nodes under $c\to 0$: it either \emph{only} moves the outer nodes, or diverges.

\subsection{Coarse-to-fine}\label{sec:schedule}
Assume thus that the gradient descent has to correct a warm start where the layout is globally wrong. Correcting it asks for rotations of relatively large angles at large radius, which only a small $c$ supplies. Once the arrangement is right the remaining work is re-arranging the inner nodes, which at $c\to 0$ no learning rate can produce without diverging the outer ones. This is precisely where the intermediate values of $c$ prove their worth, as in this case the ratio of before now becomes roughly $e^{(1-\sqrt{c})(r_{\max}-r_{\min})}$, effectively reducing the difference in movement between the outer and inner nodes. Note that the extreme case of $c=1$, which apparently cancels the radial span, would bring back the freeze.

We therefore run two stages: a coarse phase at $c\to 0$, a handover
of the coordinates, and a fine phase at an interior $c$ with a fresh optimizer state and its own learning rate.
A schedule of this kind does more than reach the same place sooner. A preconditioner leaves
the critical points of $f$ untouched, since $G_c^{-1}\nabla f$ vanishes exactly where
$\nabla f$ does, but the loss is not convex and the path decides which of them is reached.
The coarse phase is therefore not merely a faster start on one descent: it can hand over a
configuration from which the fine phase settles somewhere that the fine phase alone would
never have found. Section~\ref{sec:experiments} reports both effects against single-phase
controls run at equal budget.

\section{Experiments}
\label{sec:experiments}

\subsection{Setup and applicability screen}
In all experiments, we initialize the embeddings with the closed-form \textsc{Hydra} solution~\cite{keller2020hydra}, obtained from the eigendecomposition of $\cosh(\bbD)$, and refine it by Riemannian Adam. The pipeline carries no randomness: the warm start is a deterministic eigendecomposition and the stress is evaluated full-batch. Everything is implemented in \texttt{hypeGRL}.\footnote{Visit \url{https://github.com/CicadaUY/hypeGRL} for the library and the scripts to reproduce these experiments.}
Learning rates are searched in each graph per preconditioner curvature, and Riemannian Adam is run until convergence. Grids lie on a $1/3/10$ ladder and every reported result corresponds to the best rate in terms of loss which is \emph{bracketed}, i.e.\ it has a strictly worse or divergent neighbor on both sides; a grid whose winner fell at an end was extended until this held. We report the mean stress over the final $10\%$ of iterations.

The previous section identified a spread of radii as the necessary condition for a preconditioner to matter. We therefore screen candidate graphs by the \emph{radial span} $s = r_{\max}-r_{\min}$ of the \textsc{Hydra} warm start. Since $r_{\min}\approx 0.5$ and $r_{\max}\approx \mathrm{diam}/2$, deep unbalanced trees (of large diameter, with nodes spread over a wide range of depths) are clearly the type of graphs for which the preconditioner choice should matter, whereas for small-world graphs it should not.

The upper part of Table~\ref{tab:graphs_and_results} lists the graphs we have used in our experiments along with some statistics. We have included five trees, four of which are real phylogenies taken from the Open Tree of Life synthetic tree (release 16.1)~\cite{hinchliff2015synthesis,redelings2017supertree}: the induced subtrees of
the clades Marsupialia, Cichlidae and Falconiformes, and a rooted subtree of
Fabaceae. The fifth tree is a synthetic caterpillar: a path of $40$ nodes with $4$ leaves attached to each, which reaches a large diameter with few nodes. We have also included two hub-dominated graphs of small diameter, another regime where hyperbolic embeddings are usually applied. One is a Holme--Kim scale-free graph with tunable clustering~\cite{holme2002growing}, generated with NetworkX's \texttt{powerlaw\_cluster\_graph}~\cite{hagberg2008networkx}.
The other is the European air-traffic network of~\cite{ribeiro2017struc2vec}. 

\begin{table}[t!]
\centering
\caption{Upper part: graphs used in the experiments and some statistics. Lower part: loss (stress) for each optimization method ($\times10^3$). 
\emph{Single curvature}: one $c$
throughout, from the \textsc{Hydra} initialization. \emph{Two-stage}: $c=10^{-6}$ up to the handover (stage 1), then the $c$ in each row for stage 2. 
Gains are relative changes in loss (negative is better): curvature gain compares the best of $c\in\{0.2,0.3\}$ with $c=10^{-6}$ during stage~2; schedule gain compares the best two-stage run with the best single-curvature run. 
}
\label{tab:graphs_and_results}
\setlength{\tabcolsep}{1.5pt}
\begin{tabular}{@{}lrrrrrrr@{}}
\toprule
 & Caterp. & Marsupialia & Fabaceae & Cichlidae & Falconif. & PowerLaw & Europe \\
\midrule
$|V|$          & 200  & 837  & 1067 & 3134 & 281  & 200 & 399  \\
$|E|$          & 199  & 836  & 1066 & 3133 & 280  & 780 & 5993 \\
diameter       & 41   & 36   & 44   & 64   & 27   & 4   & 5    \\
radial span $s$& 19.9 & 17.7 & 21.8 & 27.9 & 13.2 & 1.5 & 2.4  \\
\midrule[\heavyrulewidth]
\textsc{Hydra} init. & 1763 & 6406 & 13545 & 460985 & 363 & 17.9 & 51.2 \\
\midrule
\multicolumn{8}{@{}l}{\textbf{\textit{Single curvature}}} \\
$c=10^{-6}$ & 94.4 & 137 & \textbf{1587} & \textbf{96137} & 9.02 & 14.9 & 39.7 \\
$c=0.2$     & \textbf{46.8} & \textbf{128} & 2774 & 177657 & \textbf{8.96} & 14.9 & 39.7 \\
$c=0.3$     & 287 & 943 & 3642 & 189768 & 8.98 & 14.9 & 39.7 \\
$c=1$       & 782 & 2588 & 6532 & 202851 & 92.0 & 14.9 & 39.8 \\
\midrule
\multicolumn{8}{@{}l}{\textbf{\textit{Two-stage}}} \\
At handover    & 110 & 147 & 1595 & 97916 & 9.41 & 15.0 & 39.8 \\
$c=10^{-6}$ & 73.8 & 71.1 & 863 & 67775 & 9.07 & 15.0 & 39.8 \\
$c=0.2$     & 4.32 & 35.6 & \textbf{634} & \textbf{52136} & \textbf{9.00} & 15.0 & 39.8 \\
$c=0.3$     & \textbf{0.935} & \textbf{33.2} & 681 & 63727 & 9.00 & 15.0 & 39.8 \\
\midrule
curv.\ gain (\%)  & $-98.7$ & $-53.2$ & $-26.5$ & $-23.1$ & $-0.8$ & $0.0$  & $0.0$  \\
sched.\ gain (\%) & $-98.0$ & $-74.1$ & $-60.0$ & $-45.8$ & $+0.4$ & $+0.4$ & $+0.1$ \\
\bottomrule
\end{tabular}
\end{table}
                                                 
\subsection{Single-curvature ablation}
\begin{figure}[t!]
    \centering
    \includegraphics[width=1\linewidth]{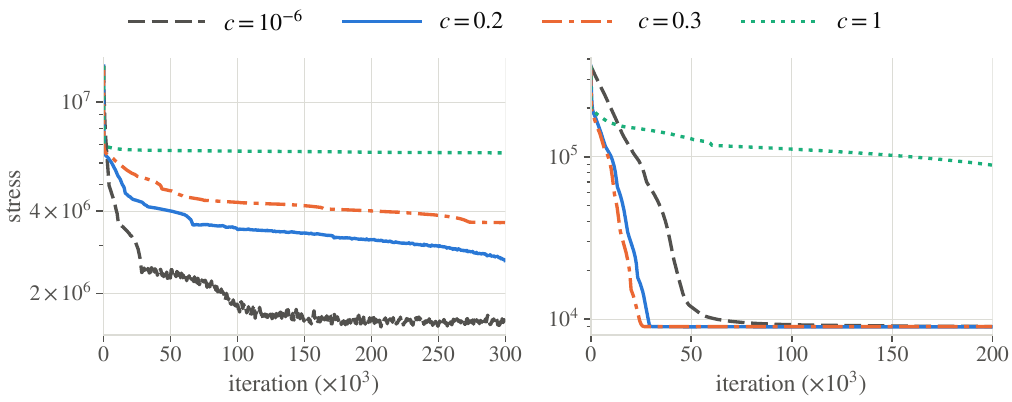}
    \caption{Stress along the iterations when a single curvature $c$ is used for the whole
optimization, starting from the \textsc{Hydra} initialization. Left: Fabaceae ($s=21.8$);
right: Falconiformes ($s=13.2$). Graphs with larger radial span $s$ are more sensitive to the pre-conditioner curvature.
}
    \label{fig:loss_single}
\end{figure}

Let us first discuss the case where we would use a single curvature throughout the whole Riemannian GD learning steps ($200$k except for Marsupialia and Fababaceae which required $300$k) after the Hydra initialization. The results are shown in the middle section of Table~\ref{tab:graphs_and_results}, and the loss curves for Fabaceae and Falconiformers in particular can be seen in Fig.~\ref{fig:loss_single}. The curves clearly illustrate that after a few iterations where the stress do decrease, the exact metric using $c=1$ is effectively frozen. Larger learning rates simply diverge, confirming the derivations we presented in the previous section.  

A second observation is that intermediate values of $c$ can produce the best results overall. This is clearly the case for graphs with smaller radial span $s$ (the last three columns in Table~\ref{tab:graphs_and_results}), where most curvatures obtain the same performance, except in the case of Falconiformes (right panel of Fig.\ \ref{fig:loss_single}). In this last case, an $s\approx 13$ is enough to freeze the GD for $c=1$. However, $c\to 0$ is systematically among the best single curvature results, so we will use it as an initialization for the two-stage scheme. The rationale is that the flat curvature rapidly rearranges the layout (see for example the left panel in Fig.~\ref{fig:loss_single}), which will then be further refined by a second stage that uses a different, intermediate curvature.

\subsection{Multi-curvature optimization}
\begin{figure}[t!]
    \centering
    \includegraphics[width=1\linewidth]{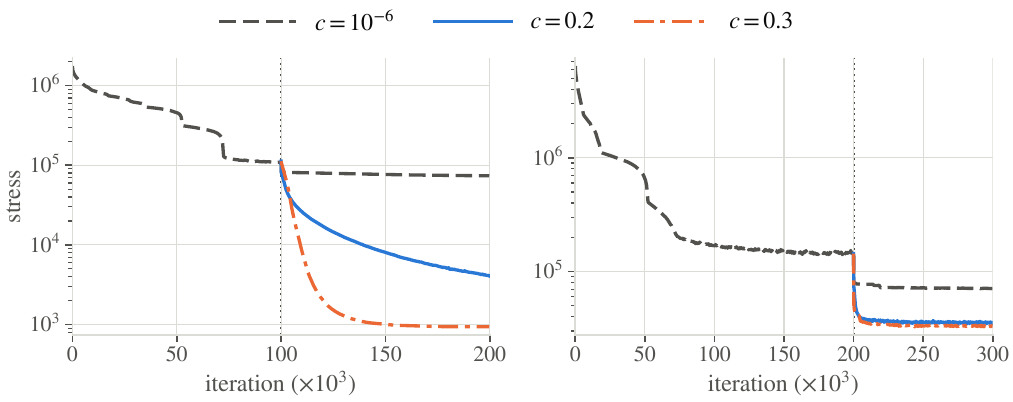}
    \caption{Stress along the iterations when the two-phase scheme is used (first $c\to 0$ and then the other curvatures), starting from the \textsc{Hydra} initialization. Left: Caterpillar;
right: Marsupialia. Even in real graphs the gain of a second phase that uses an intermediate value of $c$ is significant.
}
    \label{fig:two_phase}
\end{figure}

As we mentioned before, we now divide the learning process into two stages: after Hydra's init, we use the flat curvature $c\to 0$ (stage 1, running for $100$k steps except for Marsupialia and Fababaceae which ran for $200$k), and then restart Adam and train using another curvature (stage 2, $100$k for all graphs). In order to control the effect of simply restarting Adam, results in the lower part of Table~\ref{tab:graphs_and_results} also include the stress when using $c\to 0$ again during stage 2. The loss curves corresponding to Caterpillar and Marsupialia are shown in Fig.\ \ref{fig:two_phase}.
The first observation is that the two-phase scheme can indeed produce significantly improved results. In the toy example of Caterpillar, the stress is decreased almost by two orders of magnitude. Although in real-world graphs the gain is more modest, it is still significant, with decreases of between 25 and 50\% in the case of graphs with large radial span $s$ when compared to simply restarting Adam. These gains are far more significant when compared to choosing a single curvature throughout the whole number of iterations (see the last row of Table~\ref{tab:graphs_and_results}).  

Naturally, and as expected, in graphs with smaller radial span $s$, the gains that can be obtained from  two-stage scheme are null. The preconditioner's curvature has no effect in these graphs, visible already in the single-phase scheme, and a second phase simply does not change this fact. 

\section{Conclusions}

We have argued that, when a hyperbolic embedding is refined by gradient descent, the metric under which the steps are taken should be treated as a preconditioner, separate from the geometry of the embedding. A one-parameter family of warped metrics, indexed by a curvature $c$, spans this choice between the flat parametrization ($c\to0$) and the exact hyperbolic metric ($c=1$), while the objective stays the same function of the
embedding for every $c$. On graphs whose initialization spreads the nodes over a wide
range of radii, neither endpoint is necessarily the right choice. In fact, the exact metric, which is what standard Riemannian refinement uses, is the worst option in our experiments: from the \textsc{Hydra} initialization it reduces the stress within the first hundred
iterations and then stops making progress, ending between $2$ and $20$ times above the
best single curvature, with every larger learning rate diverging. At the other end,
continuing the refinement at $c\to0$ leaves most of the achievable reduction unused:
a second stage switching to an intermediate curvature ($c=0.2$ or $0.3$) lowers the
final stress by $23$--$53\%$ on three real phylogenies and by $98.7\%$ on the
synthetic caterpillar.

In fact, the schedule is as important as the curvature. Run from the initialization alone, an
intermediate curvature is often worse than the flat metric. Its largest advantage appears once a flat first stage has done
the large-scale rearrangement, leaving the curved metric to refine a roughly correct
configuration. At equal budget, the two-stage schedule reduces the stress by
$45$--$98\%$ relative to the best single curvature on the graphs where curvature helps.
Both effects vanish together on graphs with a small radial span. On two hub-dominated
graphs and on a phylogeny of span $13.2$, all curvatures, learning rates and schedules
reach the same stress to within $1\%$. 

\noindent\textbf{Future work.} The span acts as a threshold rather than a predictor: it
separates the graphs that benefit from those that do not, but it does not order the
size of the gain. The best curvature and learning rate were found by grid
search, separately for each graph, and a rule that sets them from the graph, or adapts
them during optimization, would make the method more practical. Finally, we have studied a
very natural but nevertheless single objective (stress); whether the same preconditioning
view carries over to other hyperbolic embedding losses, and to larger graphs, is an open question.

\section*{Funding Acknowledgements}
This work was partially funded by CSIC i+d project ``Geometr\'ia en Redes
Complejas y Aplicaciones a Aprendizaje Autom\'atico'' and Espacio Interdisciplinario of Universidad de la Rep\'ublica.

\section*{AI Use Statement}

Claude AI was used to assist in writing the manuscript and coding the simulations. The authors take full responsibility for the results of this paper.

\newpage
\bibliographystyle{plainnat}
\bibliography{refs.bib}

\end{document}